\documentclass{article}

\usepackage{microtype}
\usepackage{graphicx}
\usepackage{subcaption}
\usepackage{booktabs} 

\usepackage{hyperref}

\usepackage[accepted]{icml2026}

\usepackage{amsmath}
\usepackage{amssymb}
\usepackage{mathtools}
\usepackage{amsthm}
\usepackage{booktabs}   
\usepackage{multirow}   
\usepackage{graphicx}   
\usepackage[table]{xcolor} 

\usepackage[capitalize,noabbrev]{cleveref}

\theoremstyle{plain}

\theoremstyle{definition}

\theoremstyle{remark}

\usepackage[textsize=tiny]{todonotes}

\icmltitlerunning{Supervised Fine-Tuning Needs to Unlock the Potential of Token Priority}

\begin{document}

\twocolumn[
  \icmltitle{Supervised Fine-Tuning Needs to Unlock the Potential of Token Priority}




  \begin{icmlauthorlist}
\icmlauthor{Zhanming Shen}{yyy}
\icmlauthor{Zeyu Qin}{sch}
\icmlauthor{Jiaqi Hu}{yyy}
\icmlauthor{Wentao Ye}{yyy}
\icmlauthor{Hao Chen}{yyy}
\icmlauthor{Xiaomeng Hu}{yyy}
\icmlauthor{Haokai Xu}{yyy}
\icmlauthor{Gang Chen}{yyy}
\icmlauthor{Yi R. Fung$^\dagger$}{sch}
\icmlauthor{Haobo Wang$^\dagger$}{yyy}
\end{icmlauthorlist}

\icmlaffiliation{yyy}{Zhejiang University}
\icmlaffiliation{sch}{Hong Kong University of Science and Technology}

\icmlcorrespondingauthor{Yi R. Fung}{yrfung@ust.hk}
\icmlcorrespondingauthor{Haobo Wang}{wanghaobo@zju.edu.cn}

  \icmlkeywords{Machine Learning, ICML}

  \vskip 0.3in
]



\printAffiliationsAndNotice{}  


\begin{abstract}
The transition from fitting empirical data to achieving true human utility is fundamentally constrained by a granularity mismatch, where fine-grained autoregressive generation is often supervised by coarse or uniform signals. This position paper advocates \textbf{Token Priority} as the essential bridge, formalizing Supervised Fine-Tuning (SFT) not as simple optimization but as a precise distribution reshaping process that aligns raw data with the ideal alignment manifold. We analyze recent breakthroughs through this unified lens, categorizing them into two distinct regimes: Positive Priority for noise filtration and Signed Priority for toxic modes unlearning. We revisit existing progress and limitations, identify key challenges, and suggest directions for future research.
\end{abstract}

\section{Introduction}
\label{sec:intro}

Supervised Fine-Tuning (SFT) has established itself as the cornerstone of post-training alignment \citep{ouyang2022training,lipreserving}, effectively guiding pre-trained models toward human-centric behaviors \citep{touvron2023llama}. The prevailing paradigm treats SFT as a uniform maximum likelihood estimation problem, implicitly assuming that every token in the demonstration data contributes equally to the alignment manifold \citep{helm2025token,diao2026entropy}. We argue that this egalitarian assumption is fundamentally flawed. By treating high-entropy reasoning steps and low-entropy syntactic fillers as optimization equivalents, standard SFT fails to capture the sparse, unevenly distributed nature of intelligence \citep{fedus2022switch,ruan2025enhancing}.

In this position paper, \textbf{we argue that Supervised Fine-Tuning Needs to Unlock the Potential of Token Priority}. We posit that the transition from fitting empirical data ($P_{\text{data}}$) to achieving true general intelligence ($\pi^*$) cannot be bridged by scale alone. Instead, it requires a structural intervention: a precise, token-level reweighting mechanism that reshapes the optimization landscape. This necessity stems from the three fundamental flaws inherent in the unified SFT:

\begin{itemize}
    \item \textbf{The Information-Density Gap:} 
    The support set for general intelligence is surprisingly sparse \citep{zhou2023lima, gunasekar2023textbooks}. Indiscriminate training on broad distributions dilutes gradients with noise \citep{chen2023alpagasus}, whereas Token Priority mathematically isolates the \textit{high-density} signal required for efficient alignment, as validated by recent selection metrics \citep{caoinstruction, zhuang2025towards, pangtoken}.
    
    \item \textbf{Gradient Starvation:} 
    Optimization is structurally biased towards high-frequency \textit{imitation anchors} that monopolize the gradient budget \citep{pezeshki2021gradient, shen2026training}. Uniform SFT allows these easy tokens to drown out the rare, critical signals determining reasoning correctness, necessitating explicit reweighting to recover the tail of the distribution \citep{lin2024rho, shi2025rethinking}.
    
    \item \textbf{Exposure Bias:} 
    Teacher forcing creates a deceptive \textit{comfort zone} that blinds models to their own errors, leading to cascading hallucinations during inference \citep{bengio2015scheduled, wu2025generalization}. Priority mechanisms are essential to bridge this shift by penalizing fabricated knowledge or emphasizing recovery trajectories \citep{hu2024mitigating, shen2025don, lincritical}.
\end{itemize}

A counter-narrative suggests that these granular mismatches are temporary artifacts that will vanish with sufficient compute — the ``Scale is All You Need" hypothesis \citep{kaplan2020scaling}. We challenge this linear extrapolation. Without explicit prioritization, scaling data volume risks \textit{model collapse} \citep{shumailov2023curse} or \textit{inverse scaling}, where models overfit to spurious correlations rather than robust reasoning \citep{wei2023inverse}.

To resolve these conflicts, we introduce a unified meta-framework that formalizes post-training as a \textit{distribution reshaping} process mediated by a priority function $\Phi(x)$. We categorize recent advancements into two distinct regimes:
\begin{enumerate}
    \item \textbf{Regime I: Positive Priority (Construction).} Focusing on noise filtration and signal amplification via \textit{Hard Selection} \citep{qin2025sstoken, fu2025t} or \textit{Soft Reweighting} \citep{wang2025entropy, huang2025selectkd}.
    \item \textbf{Regime II: Signed Priority (Correction).} Bridging SFT with Reinforcement Learning by using negative gradients ($\Phi < 0$) to actively unlearn toxic modes \citep{ghahrizjani2025forgetting, hu2024mitigating}.
\end{enumerate}

We begin by formalizing the structural gap between the empirical SFT distribution and the ideal alignment manifold. Crucially, we substantiate our position by demonstrating that recent empirical breakthroughs, ranging from high-density signal extraction \citep{fu2025t, lin2024rho} to recovery-oriented training \citep{hasan2025reason2decide}, are already implicitly addressing these pathologies by reshaping optimization at the token level. Synthesizing these insights, we introduce the Token Priority Function as a unifying operator and taxonomize advancements into a unified spectrum: constructive regimes that filter noise via hard or soft selection, and corrective regimes that actively rectify hallucinations. Subsequently, we critically analyze structural bottlenecks, specifically the tension between atomic optimization and semantic integrity. We conclude by challenging the prevailing ``Scale is All You Need" hypothesis and proposing a roadmap toward dynamic, topology-aware priority schedules.


\section{A Structural Gap Between the Empirical SFT Distribution and the Ideal Alignment Manifold}
\label{sec:gap}

\subsection{Alignment as Distribution Matching Under Objective Mismatch}
\label{subsec:dist_match}

Supervised fine-tuning (SFT) of large language models (LLMs) can be viewed as searching for an \emph{ideal} conditional
policy $\pi^\star(\cdot \mid x)$ that maximizes human utility subject to logical and safety constraints. In deployment,
the object of interest is not perplexity on a static corpus, but the induced \emph{interaction} distribution over
responses that users experience:
\begin{equation}
\pi^\star \in \arg\max_{\pi \in \Pi} \; \mathbb{E}_{x \sim P_{\text{use}}}\Big[ \mathbb{E}_{y \sim \pi(\cdot \mid x)} \big[ R^\star(x,y) \big] \Big],
\label{eq:ideal_policy}
\end{equation}
where $P_{\text{use}}$ denotes the (non-stationary) query distribution at deployment, and $R^\star$ is the true (and
unobserved) human utility.

By contrast, the dominant training objectives in post-training are \emph{data-driven} surrogates. In SFT, one typically
fits an empirical instruction dataset via maximum likelihood:
\begin{equation}
\min_{\theta} \; \mathbb{E}_{(x,y)\sim P_{\text{data}}}\big[ - \log \pi_\theta(y \mid x)\big],
\label{eq:sft_mle}
\end{equation}
which implicitly assumes that the conditional label distribution in $P_{\text{data}}(y \mid x)$ is a faithful proxy for
$\pi^\star(\cdot \mid x)$. 

The central obstacle is that $P_{\text{data}}$ (and $\widehat{R}$) are not merely \emph{incomplete}; they are
\emph{structurally misaligned} with the ideal alignment manifold induced by $R^\star$ and the constraints we care about
(e.g., factuality, logical consistency, and instruction compliance). This gap is multi-dimensional, arising from
(\textbf{i}) dataset-intrinsic defects, (\textbf{ii}) distribution shift induced by training dynamics,
(\textbf{iii}) proxy failure in reward modeling, and (\textbf{iv}) fundamental limits of imitation learning.

\subsection{Information-density gap (``less is more'')}
\label{subsec:data_illusion}

A recurring empirical finding in instruction tuning is that \emph{small, carefully curated} datasets can rival or exceed
large, weakly filtered corpora, suggesting that the ``support set'' of the desired behavior is sparse.
The \emph{Superficial Alignment Hypothesis} argues that pretrained LLMs already contain most latent capabilities, and
post-training primarily selects a conversational style and a small set of behavioral modes \citep{kirstain2022few}.
Consistent with this view, LIMA reports strong alignment with only $\mathcal{O}(10^3)$ high-quality examples
\citep{zhou2023lima}. Similarly, automated quality scoring and filtering improves performance by
discarding a large fraction of instruction data \citep{tunstall2023zephyr,zhang2024recost,chen2023alpagasus,zhao2024long,park2025instruct}. These results highlight an
\emph{information-density gap}: large portions of $P_{\text{data}}$ place probability mass on low-signal or even
counterproductive regions of the space, diluting gradients away from the ideal manifold.

\subsection{Gradient Starvation}
\label{Gradient_Starvation}
Even if $P_{\text{data}}$ were clean, learning objectives can amplify \emph{spurious} features correlated with
perceived quality \citep{gui2025mitigating,chen2025safety}. Standard SFT minimizes the negative log-likelihood uniformly across all tokens: $\mathcal{L} = -\sum_t \log \pi(y_t|s_t)$.
However, natural language follows a heavy-tailed distribution where syntax and common phrases (``anchors'') appear orders of magnitude more frequently than semantic pivot points \citep{li2025steering,zimmerman2024tokens}.
This leads to \textbf{gradient starvation} \citep{pezeshki2021gradient,ludan2023explanation}:
\begin{equation}
\mathbb{E}\big[\|\nabla_\theta \mathcal{L}\| \,\big|\, y_t \in \mathcal{T}_{\text{easy}}\big] \gg \mathbb{E}\big[\|\nabla_\theta \mathcal{L}\| \,\big|\, y_t \in \mathcal{T}_{\text{hard}}\big].
\label{eq:rare_grad}
\end{equation}
The optimization budget is exhausted on fitting easy tokens (driving their loss to near-zero) before the model can learn the rare, robust features required for generalization \citep{liu2025ea4llm}. What appears as a capability failure is often a training-dynamics artifact: the critical tokens were simply ``drowned out'' by the noise of the easy ones \citep{lincritical}.



\subsection{Exposure Bias}
\label{subsec:exposure_bias}

The structural gap between empirical data and general intelligence is perhaps most mathematically acute in the phenomenon of \textit{Exposure Bias} \citep{bengio2015scheduled,arora2022exposure,xu2019rethinking}. While often treated as a mere variance reduction technique, we argue that Teacher Forcing fundamentally alters the state distribution, creating a mismatch that uniform SFT cannot bridge \citep{ross2011reduction,lin2024rho}.

\textbf{The Mathematical Essence of Covariate Shift.}
Standard autoregressive training minimizes the negative log-likelihood of the target token $x_t$ conditioned on the \textit{ground truth} history $x^*_{<t}$:
\begin{equation}
    \mathcal{L}_{\text{SFT}}(\theta) = \mathbb{E}_{x \sim \mathcal{D}} \left[ -\sum_{t=1}^{T} \log \pi_\theta(x_t \mid x^*_{<t}) \right].
\end{equation}
This objective implicitly assumes that the state distribution encountered during inference, $d_{\pi_\theta}(s)$, is isomorphic to the training distribution $d_{\pi_{\text{data}}}(s)$ \citep{ranzato2015sequence}. However, during inference, the model must condition on its \textit{own} generated history $\hat{x}_{<t}$. Since $\pi_\theta$ cannot perfectly fit $\pi_{\text{data}}$, microscopic errors $\epsilon$ at step $t$ accumulate, causing a deviation in the trajectory that grows with sequence length $T$---a phenomenon known as \textbf{Cascading Errors} \citep{ross2010efficient,wu2025survey}.

This is formally characterized in statistical learning theory as \textbf{Covariate Shift} \citep{shimodaira2000improving,hasan2025reason2decide}. Specifically, while the conditional probability $\pi(y|x)$ (the optimal action given a state) might be invariant, the input marginal distribution $P(x)$ (the state space visited) shifts drastically. Recent work within the Imitation Learning framework highlights that this mismatch results in a super-linear growth of the error bound, driving the model into "undefined regions" (Out-of-Distribution, OOD) \cite{kumar2020conservative,foster2024behavior}. In these regions, lacking support from the training data, the model's behavior inevitably degenerates into hallucination.

\textbf{The Comfort Zone Trap of Teacher Forcing.}
Beyond distribution shift, Teacher Forcing constructs a deceptive "comfort zone" \citep{jiang2020can,shen2026training}. During training, regardless of how erroneous the model's prediction $\hat{x}_t$ is, the input at $t+1$ is reset to the ground truth $x^*_t$. This mechanism effectively severs the causal link between error propagation and future states, instilling a false sense of overconfidence.

Consequently, uniform SFT creates a critical blind spot: \textbf{the model never learns to recover from failure} \citep{holtzmancurious,hasan2025reason2decide}. In an ideal general intelligence system, reasoning is a dynamic process of self-correction. If a Chain-of-Thought (CoT) trajectory deviates, the model should possess the capability to generate ``Recovery Tokens'' that steer the trajectory back onto the correct logical manifold \citep{wang2025beyond,lincritical}. However, SFT treats all training samples as expert trajectories, implicitly signaling that "all preceding steps are correct." Thus, when the model inevitably drifts during inference, it fails to recognize the error and instead conditions on the flawed context to continue reasoning, leading to total logical collapse \cite{lincritical,allenphysics}.

\section{Bridging the Gap: Why Alignment Must Be Token-Level}
\label{sec:bridging_the_gap}

In Section~\ref{sec:gap}, we identified three structural failures in current post-training: the redundancy of empirical data (data illusion), the tyranny of easy tokens (gradient starvation), and the comfort zone of teacher forcing (exposure bias). We argue that these are not separate artifacts but symptoms of a single root cause: \textit{granularity mismatch}. The transition from $P_{\text{data}}$ to $\pi^\star$ cannot be achieved via scalar rewards or uniform likelihoods; it requires explicitly reshaping the optimization landscape at the token level.

\subsection{Operationalizing the Superficial Alignment Hypothesis: From Filtering to Sparsity}
\label{subsec:operationalizing_alignment}

The \textit{Data Illusion} (Section~\ref{subsec:data_illusion}) suggests that the vast majority of post-training tokens are redundant. Token Priority allows us to operationalize the \textit{Superficial Alignment Hypothesis} by shifting the paradigm from "learning new knowledge" to "sparse capability elicitation."

\textbf{High-Density Signal Extraction.}
If alignment is merely mode selection, then training on low-information tokens is not just inefficient—it is harmful. The future lies in \textbf{Token-Level Distillation}, where we mathematically isolate the "support set" of alignment.
Recent works validate this by pruning datasets not at the sample level, but at the token level. Approaches like \textbf{T-Shirt} \citep{fu2025t}, \textbf{Token Cleaning} \citep{pangtoken}, and \textbf{SelectIT} \citep{liu2024selectit} demonstrate that masking uninformative regions based on uncertainty or hierarchy yields superior generalizability. Similarly, \textbf{InstructMining} \citep{caoinstruction} and \textbf{DIQ} \citep{zhuang2025towards} prove that selecting samples based on "learning difficulty" rather than surface quality prevents the dilution of gradients.
This trend suggests a move towards \textbf{Information-Theoretic Training}, where the loss function $\mathcal{L}$ is computed exclusively on tokens that provide non-zero information gain relative to the pre-trained prior \citep{lin2024rho,liu2026profit}.

\textbf{Structural Sparsity and Knowledge Integrity.}
The "Less is More" principle also extends to the model's parameter space.
To avoid the "alignment tax" (forgetting or hallucination), we must enforce \textbf{Structural Sparsity}.
Methods like \textbf{S2FT} \citep{yang2024s} and \textbf{SFTKey} \citep{shi2025rethinking} show that updating only a fraction of parameters or focusing solely on "Key" tokens preserves the model's pre-trained world model while adapting its style.
Furthermore, preserving the "Knowledge Boundary" is critical; techniques like \textbf{SEAT} \citep{shen2025don} and \textbf{Forgetting} \citep{ghahrizjani2025forgetting} explicitly protect the model's "Ignorance Awareness," ensuring that SFT does not force the model to hallucinate on unknown concepts.
By combining faithful fine-tuning objectives \citep{chaduvula2026reducing,hu2025fine} with attention-guided token selection \citep{arif2025paint}, we can achieve alignment that is both data-efficient and robust to the noise inherent in $P_{\text{data}}$ \citep{luo2024robustft}.

\subsection{Redistributing the Optimization Budget: Solving Gradient Starvation}
\label{subsec:solving_starvation}

In Section~\ref{Gradient_Starvation}, we identified \textit{Gradient Starvation} as a systemic failure where the finite optimization budget is monopolized by high-frequency ``easy'' tokens (anchors), leaving complex reasoning features under-learned. Token Priority offers a principled solution by acting as a \textbf{Dynamic Gradient Equalizer}.

\textbf{The Economics of Gradient Allocation.}
Standard SFT allocates gradient updates proportional to token frequency, implicitly assuming that the most frequent patterns are the most valuable. Priority-based training challenges this by treating gradient capacity as a scarce economic resource that must be redistributed from ``redundant'' to ``informative'' regions.
By assigning priority weights $\Phi(x_t)$ inversely proportional to a token's triviality, we can mathematically enforce \textit{Gradient Equity}:
\begin{equation}
    \mathbb{E}\big[\Phi_{\text{easy}} \cdot \|\nabla_\theta \mathcal{L}\|\big] \approx \mathbb{E}\big[\Phi_{\text{hard}} \cdot \|\nabla_\theta \mathcal{L}\|\big].
\end{equation}
Recent works implement this redistribution through two complementary mechanisms: \\
\textbf{(1) Suppressing the Trivial:} Methods such as \textbf{Rho-1} \citep{lin2024rho}, \textbf{EntroDrop} \citep{wang2025entropy}, and \textbf{T-Shirt} \citep{fu2025t} effectively ``freeze'' parameters on low-loss, high-frequency patterns. By pruning or down-weighting these easy tokens \citep{liu2024selectit, pangtoken}, they prevent the optimizer from wasting updates on surface-level fluency that the model has already mastered. \\
\textbf{(2) Amplifying the Critical:} Conversely, strategies like \textbf{ssToken} \citep{qin2025sstoken}, \textbf{ProFit} \citep{liu2026profit}, and \textbf{GIFT} \citep{xu2026giftguidedimportanceawarefinetuning} identify sparse, high-information signals---whether defined by high entropy, semantic contribution, or training dynamics---and amplify their gradient contribution. This ensures that the ``long-tail'' of reasoning pivots receives sufficient signal to drive generalization, even when these tokens are statistically rare in the corpus.

\textbf{Escaping the ``Basin of Ease''.}
This redistribution is essential for escaping the ``Basin of Ease''---a local minimum where loss is low due to syntactic mastery, but reasoning remains fragile.
Nuanced approaches \citep{huang2025low,yu2025adaptive} further refine this by distinguishing between \textit{destabilizing noise} and \textit{valuable exploration sparks}, ensuring that budget cuts do not inadvertently suppress necessary uncertainty.
Ultimately, Token Priority transforms the training process from \textit{fitting data density} to \textit{maximizing information acquisition} \citep{lei2025revisiting}, suggesting that future scaling laws will depend not on raw token count, but on \textbf{Priority-Weighted Token Count}.

\subsection{From Exposure Bias to Error Recovery: The Distributional Repair}
\label{subsec:error_recovery}

In Section~\ref{subsec:exposure_bias}, we identified \textit{Exposure Bias} not merely as error accumulation, but as a fundamental state-distribution mismatch where models trained in the "comfort zone" of teacher forcing fail to learn a \textit{recovery policy}. Token Priority provides the necessary mechanism to bridge this gap by reshaping the training distribution to resemble the operational manifold.

\textbf{Training for Recovery, Not Just Imitation.}
To prevent the logical collapse described in Section~\ref{subsec:exposure_bias}, training must explicitly value tokens that demonstrate \textit{corrective capability}. While standard SFT treats all ground-truth tokens as equal, a Priority-based framework can dynamically up-weight "Recovery Tokens"—those that resolve ambiguity or correct deviations in non-expert trajectories. Recent work on task-level scheduled sampling \cite{hasan2025reason2decide} exemplifies this principle: by forcing the model to rationalize its own predictions, we implicitly assign high priority to tokens that function effectively outside the gold-standard distribution. This transforms the objective from $\max P(y|x_{\text{gold}})$ to maximizing the probability of reaching the solution manifold from a perturbed state $\tilde{x}$.

\textbf{Decoupling Local Validity from Global Correctness.}
A rigid sequence-level view discards valuable signals from imperfect trajectories. As observed in \textbf{Rho-1} \cite{lin2024rho} and \textbf{Critical Token Selection} \cite{lincritical}, a reasoning chain is not a monolith; it contains both "noise" and "sparks." Token Priority allows us to decouple these signals: \\
\textbf{(1) Filtering Noise in SFT:} By identifying and masking easy, surface-level patterns \cite{shen2026training}, we prevent the model from overfitting to the static training distribution, forcing it to focus on the "long-tail" logic tokens that dictate generalization. \\
\textbf{(2) Mining Value in Failure:} Conversely, even in incorrect trajectories, specific transition steps may represent valid logical operations. A fine-grained Priority function $\Phi(x_t)$ can reward these locally optimal actions via selective backpropagation, ensuring that the model learns robust logical operators rather than just memorizing successful paths.

\textbf{Reshaping Conditional Probability via Importance Sampling.}
Mathematically, correcting the discrepancy between the training distribution $\pi_{\text{data}}$ and the optimal inference distribution $\pi^*$ requires importance sampling. Methods like \textbf{TIS-DPO} \cite{liutis} demonstrate that Token Priority can act as the density ratio $w_t = \frac{\pi^*(x_t)}{\pi_{\text{ref}}(x_t)}$. By integrating these weights into the loss, we effectively simulate the inference-time distribution during training, ensuring that gradients are concentrated precisely where the model's operational policy diverges from the ideal logic, rather than uniformly across the sequence.



\section{The Token Priority Spectrum: A Unified Framework}
\label{sec:framework}

Existing post-training methodologies are often fragmented into discrete tasks—filtering for SFT, credit assignment for RL, and re-weighting for distillation. We argue that these are not separate problems but rather different approximations of a single mathematical objective: the density reshaping of the training distribution. In this section, we present a unified meta-framework that formalizes \textit{Token Priority} as the fundamental operator bridging the gap between empirical data and general intelligence.

\subsection{The Distributional Gap and The Priority Function}

Let $\mathcal{P}_{\text{data}}$ denote the empirical distribution of the available training corpus (e.g., web text, trajectories). Let $\mathcal{P}_{\text{ideal}}$ denote the theoretical distribution of an optimal agent—one that is factually accurate, logically rigorous, and safely aligned.
The fundamental axiom of our position is that $\mathcal{P}_{\text{data}} \neq \mathcal{P}_{\text{ideal}}$. Standard uniform training implicitly assumes $\mathcal{P}_{\text{data}} \approx \mathcal{P}_{\text{ideal}}$, treating every token as an equal contributor to the target manifold. This assumption is flawed due to noise, redundancy, and toxicity inherent in raw distributions.

To bridge this gap, we define a continuous mapping operator, the \textbf{Token Priority Function} $\Phi: \mathcal{V} \times \mathcal{C} \to \mathbb{R}$, which assigns a scalar value to a token $x_t$ given context $x_{<t}$. The relationship between the distributions is governed by:
\begin{equation}
    \mathcal{P}_{\text{ideal}}(x) \propto \Phi(x) \cdot \mathcal{P}_{\text{data}}(x)
\end{equation}
Here, $\Phi(x)$ acts as a density ratio estimator (akin to the Radon-Nikodym derivative). The training objective transforms from minimizing negative log-likelihood to a priority-weighted optimization:
\begin{equation}
    \mathcal{L}(\theta) = -\mathbb{E}_{x \sim \mathcal{P}_{\text{data}}} \left[ \sum_{t} \Phi(x_t | x_{<t}) \cdot \log \pi_\theta(x_t | x_{<t}) \right]
\end{equation}

\subsection{The Geometry of Priority}

The mathematical nature of the transformation from $\mathcal{P}_{\text{data}}$ to $\mathcal{P}_{\text{ideal}}$ depends entirely on the codomain (value range) of $\Phi$. We identify three critical zones in the \textit{Token Priority Spectrum}:

\begin{itemize}
    \item \textbf{Zone I: The Constructive ($\Phi > 0$)}. Tokens in this range contribute positively to the ideal distribution. The magnitude represents the \textit{information gain} or \textit{learning necessity}. A value of $\Phi \gg 1$ implies the token is a sparse, critical signal (e.g., a logic cutpoint), while $0 < \Phi < 1$ implies high redundancy (e.g., common syntax).
    \item \textbf{Zone II: The Neutral ($\Phi = 0$)}. Tokens that carry no marginal utility for the target task. Training on them is computationally wasteful but not necessarily harmful.
    \item \textbf{Zone III: The Destructive ($\Phi < 0$)}. Tokens that actively diverge from $\mathcal{P}_{\text{ideal}}$ (e.g., hallucinations, toxic content, or broken reasoning). Optimizing for these tokens moves the model parameters \textit{away} from the ideal manifold.
\end{itemize}

This spectrum necessitates two distinct meta-strategies for approximating $\Phi$, depending on whether we view the ideal distribution as a \textit{subset} or a \textit{deformation} of the data.

\subsection{Meta-Strategy A: Soft Reweighting}
\textit{Assumption: $\mathcal{P}_{\text{ideal}}$ is a weighted deformation of $\mathcal{P}_{\text{data}}$.}

This paradigm posits that "correctness" is not binary and that gradient updates should be scaled continuously by the token's importance. By treating the priority function as a continuous spectrum, Soft Reweighting theoretically offers a \textbf{higher performance ceiling}, as it can recover the fine-grained geometry of the ideal manifold by amplifying scarce signals ($\Phi > 1$) and effectively suppressing easy ones ($0 < \Phi < 1$).
\begin{itemize}
    \item \textbf{Mechanism:} $\Phi$ is continuous, i.e., $\Phi \in \mathbb{R}$.
    \item \textbf{The Precision Requirement:} The cost of this higher ceiling is fragility. This strategy requires a \textit{precise estimation} of the target distribution; if the priority proxy is noisy, continuous scaling can amplify artifacts, destabilizing the training dynamics.
    \item \textbf{Role of Negative Values:} This framework can explicitly handle $\Phi < 0$ by applying \textit{gradient reversal} or \textit{unlearning} objectives, actively pushing the model away from harmful regions rather than just ignoring them.
\end{itemize}

\subsection{Meta-Strategy B: Hard Selection}
\textit{Assumption: $\mathcal{P}_{\text{ideal}} \subset \mathcal{P}_{\text{data}}$.}

In contrast, Hard Selection assumes that the ideal distribution is already "buried" within the noisy data. It collapses the infinite complexity of the priority spectrum into a binary decision boundary: Keep or Drop. While this is a \textbf{coarser approximation} that loses the nuance of signal magnitude, it is significantly \textbf{more robust} to estimation errors.
\begin{itemize}
    \item \textbf{Mechanism:} $\Phi$ is binary, i.e., $\Phi \in \{0, 1\}$.
    \item \textbf{The Robustness Trade-off:} By thresholding the signal, this method acts as a stable high-pass filter. It sacrifices the ability to modulate learning rates for specific tokens in exchange for stability, ensuring that the model is protected from low-confidence noise even when the priority estimation is imperfect.
    \item \textbf{Role of Negative Values:} Under this view, if a token is harmful ($\Phi < 0$ in theory), it is treated simply as "not selected" ($\Phi = 0$).
\end{itemize}

This meta-framework reveals that the choice between Hard Selection and Soft Reweighting is not merely algorithmic preference, but a fundamental stance on the nature of the training data and its relationship to the desired intelligence.

\begin{table*}[t]
\caption{\textbf{The Priority Spectrum:} A taxonomy of methods sorted chronologically (implicit) within each regime. The \textbf{Signal Source} column distinguishes between methods deriving priority from internal model states (Intrinsic) versus external supervision (External). The \textbf{Definition of $\Phi(x)$} illustrates the mathematical construction.}
\label{tab:token_priority_range}
\centering
\scriptsize
\resizebox{\textwidth}{!}{%
\begin{tabular}{@{}l l l l l p{5.5cm}@{}}
\toprule
\textbf{Regime} & \textbf{Method} & \textbf{Proxy Signal} & \textbf{Signal Source} & \textbf{Definition of Priority Function $\Phi(x)$} \\
\midrule

\multicolumn{5}{c}{\cellcolor{gray!15}\textbf{\textit{Regime I: Positive Priority ($\Phi \ge 0$) — ``Constructive Selection"}}} \\
\midrule

\multirow{15}{*}{\shortstack[l]{\textbf{Hard Selection}\\ ($\Phi \in \{0, 1\}$)}}
 & \textbf{Rho-1} \citep{lin2024rho} & Loss Gap & Intrinsic & $\Phi(x) = \mathbb{I}[\mathcal{L}_{\theta}(x) < \mathcal{L}_{ref}(x)]$ \\
 & \textbf{S2FT} \citep{yang2024s} & Gradient Sensitivity & Intrinsic & $\Phi(x_{head}) = \mathbb{I}[\nabla_{W} \mathcal{L} > \tau]$, selects sparse heads. \\
 & \textbf{Instruction Mining} \citep{caoinstruction} & Data Influence & Intrinsic & $\Phi(S) = \mathbb{I}[\text{Infl}(S, D_{val}) > 0]$, selects helpful subsets. \\
 & \textbf{Faithful FT} \citep{hu2024mitigating} & Factuality Label & External & $\Phi(x) = \mathbb{I}[x \text{ is Factual}]$, masks hallucinations. \\
 
 & \textbf{SFTKey} \citep{shi2025rethinking} & Semantic Role & External & $\Phi(x) = \mathbb{I}[x \text{ is KeyAnswer}]$, masks CoT/Prompt. \\
 & \textbf{ssToken} \citep{qin2025sstoken} & Training Dynamics & Intrinsic & $\Phi(x) = \mathbb{I}[x \in \text{Top-}\rho(\text{REL} + \text{Attn})]$ \\
 & \textbf{Token Cleaning} \citep{pangtoken} & Memorization & Intrinsic & $\Phi(x) = \mathbb{I}[\text{Infl}(x) \text{ indicates not memorized}]$ \\
 & \textbf{SelectIT} \citep{liu2024selectit} & Self-Reflection & Intrinsic & $\Phi(x) = \mathbb{I}[\text{Confidence}(x) \text{ via Self-Reflection} > \tau]$ \\
 & \textbf{T-Shirt} \citep{fu2025t} & Info Gain & Intrinsic & $\Phi(x) = \mathbb{I}[\text{IG}(x) > \tau_{tok}] \cdot \mathbb{I}[\text{IG}(Chunk) > \tau_{chk}]$ \\
 & \textbf{OpenCoder} \citep{huang2025opencoder} & Code Quality & External & $\Phi(x) = \mathbb{I}[\text{Passes Heuristics} \land \text{Classifier} > \tau]$ \\
 & \textbf{Critical Token} \citep{lincritical} & Counterfactual Impact & Intrinsic & $\Phi(x) = \mathbb{I}[|\mathcal{L}(y) - \mathcal{L}(y_{\setminus x})| > \delta]$ \\
 & \textbf{DIQ} \citep{zhuang2025towards} & Diff-Influence Map & Intrinsic & $\Phi(S) = \mathbb{I}[S \in \text{HighDiff} \cap \text{HighInfl}]$ \\
 & \textbf{Anchored SFT} \citep{zhu2025anchored} & Anchor Distance & Intrinsic & $\Phi(x) = \mathbb{I}[D_{KL}(P_{\theta}||P_{anchor}) < \delta]$ \\
 
 & \textbf{T3S} \citep{shen2026training} & Learning Speed & Intrinsic & $\Phi(x) = \mathbb{I}[x \notin \text{ImitationAnchors} (\text{Learned Early})]$ \\
\midrule

\multirow{10}{*}{\shortstack[l]{\textbf{Soft Reweighting}\\ ($\Phi \in \mathbb{R}^+$)}}
 & \textbf{RobustFT} \citep{luo2024robustft} & Noise Est. & Intrinsic & $\Phi(x) = \exp(-\gamma \cdot \text{EstimatedNoise}(x))$ \\
 & \textbf{WIT} \citep{chatterjee2025effect} & Token Role & External & $\Phi(x) = w_{resp} \mathbb{I}_{resp} + w_{prmpt} \mathbb{I}_{prmpt}$ (Step Function) \\
 
 & \textbf{EntroDrop} \citep{wang2025entropy} & Token Entropy & Intrinsic & $\Phi(x) \sim \text{Bernoulli}(1 - e^{-\lambda H(x)})$ \\
 & \textbf{SelecTKD} \citep{huang2025selectkd} & Teacher Conf. & Intrinsic & $\Phi(x) = P_{teacher}(x | context)$ \\
 & \textbf{SRFT} \citep{fu2025srft} & Entropy Gap & Intrinsic & $\Phi(x) \propto \exp(-|H_{SFT} - H_{RL}|)$, aligns policies. \\
 & \textbf{DFT} \citep{wu2025generalization} & Model Prob. & Intrinsic & $\Phi(x) = 1 - P_{\theta}(x)$ (Inverse Probability Weighting) \\
 & \textbf{ATDP} \citep{yu2025adaptive} & Utility Score & External & $\Phi(x) \propto \text{Utility}(x) / \text{PrivacyBudget}$ \\
 & \textbf{PAINT} \citep{arif2025paint} & Visual Attn. & External & $\Phi(x) = 1 + \alpha \cdot \text{CrossAttn}(Image, x)$ \\
 & \textbf{SWIFT} \citep{letoken} & Self-Play Gap & Intrinsic & $\Phi(x) = \sigma(P_{\theta}(x) - P_{ref}(x))$ \\
 & \textbf{EAFT} \citep{diao2026entropy} & Confident Conflict & Intrinsic & $\Phi(x) = \text{Gate}(H(x))$, suppresses confident errors. \\
\midrule

\multicolumn{5}{c}{\cellcolor{red!10}\textbf{\textit{Regime II: Signed Priority ($\Phi \in \mathbb{R}$) — ``Corrective Rectification"}}} \\
\midrule

\multirow{4}{*}{\shortstack[l]{\textbf{Signed /}\\ \textbf{Unlearning}\\ ($\exists \Phi < 0$)}}
 & \textbf{Forgetting} \citep{ghahrizjani2025forgetting} & Negative Label & External & $\Phi(x_{neg}) = -1$ (Gradient Ascent) \\
 & \textbf{Ignorance Aware} \citep{shen2025don} & Unknown Status & External & $\Phi(x) = -1$ if $x$ is "made up" knowledge. \\
 & \textbf{Factuality Pref.} \citep{hu2025fine} & Hallucination & External & $\Phi(x) = \text{DPO}(x_{factual} > x_{hallucinated})$ \\
 & \textbf{F-DPO} \citep{chaduvula2026reducing} & Truthfulness & External & $\Phi(x) = \text{DPO Pair} + \text{FactualityMargin}$ \\
\bottomrule
\end{tabular}
}
\vspace{-1em}
\end{table*}

\section{Methodological Instantiations}
\label{sec:instantiations}

Current methodologies can be rigorously taxonomized by the \textit{codomain} of their priority function $\Phi$. This distinction determines not just how the model learns, but how it treats the "negative space" of the data distribution—whether errors are simply ignored (Zeroed) or actively penalized (Negated). A comprehensive overview of these methodologies is synthesized in \textbf{Table~\ref{tab:token_priority_range}}.

\subsection{Regime I: Positive Priority ($\Phi \ge 0$) — The Logic of Construction}
\textit{Assumption: $\mathcal{P}_{\text{ideal}}$ is extracted from $\mathcal{P}_{\text{data}}$. Negative data is noise to be ignored.}

This regime dominates standard Supervised Fine-Tuning (SFT). The optimization objective is strictly constructive: maximize likelihood on selected regions. "Bad" tokens are treated as having zero utility ($\Phi=0$) or reduced weight ($\Phi < 1$), effectively modulating their contribution to the gradient calculation.

\textbf{Hard Selection ($\Phi \in \{0, 1\}$): The High-Pass Filter.} 
The most common instantiation collapses the spectrum into a binary mask ($\mathbb{I}[\cdot]$), retaining only tokens that satisfy a specific quality threshold.
\begin{itemize}
    \item \textbf{Metric-Based Filtration:} Early approaches focused on static metrics. \textbf{Rho-1} \citep{lin2024rho} utilizes the model-intrinsic \textit{Loss Gap} between a target and reference model, defining $\Phi(x) = \mathbb{I}[\mathcal{L}_{\theta} < \mathcal{L}_{ref}]$. Conversely, \textbf{Faithful FT} \citep{hu2024mitigating} relies on external \textit{Factuality Labels} to mask non-factual tokens ($\Phi=0$) during training. \textbf{Instruction Mining} \citep{caoinstruction} calculates \textit{Data Influence} on a validation set to select helpful subsets while discarding negative transfer data.
    
    \item \textbf{Structural \& Semantic Priors:} Selection increasingly targets specific model components or semantic roles. \textbf{S2FT} \citep{yang2024s} employs intrinsic \textit{Gradient Sensitivity} to select sparse attention heads ($\mathbb{I}[\nabla > \tau]$) for efficient updates. In reasoning, \textbf{SFTKey} \citep{shi2025rethinking} utilizes external \textit{Semantic Roles}, setting $\Phi=1$ only for "Key Answer" tokens while masking Chain-of-Thought based on structural tags. \textbf{T-Shirt} \citep{fu2025t} extends this to \textit{Info Gain}, hierarchically selecting tokens and chunks that maximize information density. \textbf{Critical Token} \citep{lincritical} refines this by identifying \textit{Counterfactual Impact}, selecting only those tokens whose removal significantly alters the loss.
    
    \item \textbf{Dynamics \& Self-Evolution:} Recent methods leverage the training trajectory itself. \textbf{ssToken} \citep{qin2025sstoken} and \textbf{Token Cleaning} \citep{pangtoken} monitor \textit{Training Dynamics} and \textit{Memorization}, selecting tokens that are difficult but learnable while discarding overfitted ones. \textbf{SelectIT} \citep{liu2024selectit} employs \textit{Self-Reflection} to filter samples based on model confidence. \textbf{OpenCoder} \citep{huang2025opencoder} and \textbf{DIQ} \citep{zhuang2025towards} combine these insights with external domain-specific \textit{Code Quality} rules or \textit{Diff-Influence Maps}. \textbf{Anchored SFT} \citep{zhu2025anchored} introduces \textit{Anchor Distance} constraints. Most recently, \textbf{T3S} \citep{shen2026training} identifies intrinsic \textit{Learning Speed} to mask "Imitation Anchors"—easy tokens that are learned too early and hinder generalization.
\end{itemize}

\textbf{Soft Reweighting ($\Phi \in \mathbb{R}^+$): The Modulation.} 
Methods in this category assign continuous positive weights to modulate learning speed without severing dependencies or discarding data entirely.
\begin{itemize}
    \item \textbf{Noise \& Role Modulation:} \textbf{RobustFT} \citep{luo2024robustft} tackles noisy data by down-weighting samples based on \textit{Noise Estimation} ($\Phi \propto \exp(-\text{Noise})$). \textbf{WIT} \citep{chatterjee2025effect} applies step-function weights based on external \textit{Token Role}, assigning different scalar values to prompts versus responses.
    
    \item \textbf{Entropy, Confidence \& Privacy:} As models mature, weighting becomes more adaptive via intrinsic signals. \textbf{EntroDrop} \citep{wang2025entropy} modulates dropout rates using \textit{Token Entropy}, while \textbf{SelecTKD} \citep{huang2025selectkd} weights distillation loss by \textit{Teacher Confidence} ($P_{teacher}$). A standout approach in this domain is \textbf{EAFT} \citep{diao2026entropy}, which addresses ``Confident Conflicts''—tokens where the model is certain yet incorrect. By utilizing an intrinsic entropy gate to suppress these destructive updates, EAFT offers a computationally elegant, reference-free solution to mitigate catastrophic forgetting. In broader applications, \textbf{ATDP} \citep{yu2025adaptive} allocates differential privacy budgets based on external \textit{Utility Scores} derived from PII detection. In multimodal contexts, \textbf{PAINT} \citep{arif2025paint} leverages external grounding signals, up-weighting tokens with high \textit{Visual Attention} to improve fidelity.
    
    \item \textbf{Policy Alignment:} Advanced reweighting aligns SFT with RL principles. \textbf{SRFT} \citep{fu2025srft} weights updates by the \textit{Entropy Gap} between policies, and \textbf{DFT} \citep{wu2025generalization} uses inverse \textit{Model Probability} to rectify the reward implicit in SFT. \textbf{SWIFT} \citep{letoken} extends this to self-play, defining $\Phi$ via the \textit{Self-Play Gap} ($\sigma(P_\theta - P_{ref})$) to implicitly optimize preferences.
\end{itemize}

\subsection{Regime II: Signed Priority ($\Phi \in \mathbb{R}$) — The Logic of Correction}
\textit{Assumption: $\mathcal{P}_{\text{ideal}}$ requires suppressing $\mathcal{P}_{\text{toxic}}$. Negative data is a repulsive signal.}

This regime bridges SFT with Reinforcement Learning. It acknowledges that simply ignoring hallucinations is insufficient; the model must be explicitly pushed away from them ($\Phi < 0$).

\begin{itemize}
    \item \textbf{Negative Rectification \& Unlearning:} \textbf{Forgetting Mechanisms} \citep{ghahrizjani2025forgetting} employ gradient ascent ($\Phi = -1$) on samples with external \textit{Negative Labels} to erase harmful behaviors. \textbf{Ignorance Awareness} \citep{shen2025don} specifically targets \textit{Unknown Status}, penalizing the model when it attempts to "make up" answers to known-unknown questions.
    
    \item \textbf{Epistemic Correction:} To combat hallucinations via preference learning, \textbf{Factuality Preference} \citep{hu2025fine} and \textbf{F-DPO} \citep{chaduvula2026reducing} construct signed priority functions using DPO objectives. They effectively assign negative gradients to non-factual tokens relative to factual ones, using external \textit{Hallucination} status or \textit{Truthfulness} margins to reshape the probability landscape away from fabrication.
\end{itemize}

\subsection{The Signal Source: Intrinsic vs. External}
A critical dimension of our taxonomy, as highlighted in the "Signal Source" column of Table~\ref{tab:token_priority_range}, is the origin of the priority signal. Methods typically rely on either \textbf{Model-Intrinsic} signals or \textbf{External} signals:

\textbf{Model-Intrinsic Signals} (e.g., Entropy, Loss Gap, Gradient Sensitivity) exploit the model's internal dynamics to estimate data utility. These signals offer high scalability and enable self-evolution, as they require no expensive annotations. However, they are susceptible to a "blind leading the blind" failure mode, where the model reinforces confident hallucinations simply because they exhibit low entropy or high internal consistency (see Appendix~\ref{sec:blind}).

\textbf{External Signals} (e.g., Factuality Labels, Visual Attention, Code Execution) ground the priority function in outside truth or multimodal constraints. This breaks the circularity of self-training and provides necessary grounding. The trade-off is often higher curation cost, domain dependency, or the need for oracle supervision.

\section{From Limitations to Challenges}
\label{sec:challenges}

While the Token Priority framework offers a unified theoretical lens, its practical realization faces three critical bottlenecks detailed in Appendix~\ref{app:challenges}. First, the \textbf{Granularity Mismatch} between atomic optimization and semantic integrity risks severing dependencies, necessitating a shift from scalar weighting to \textit{Topological Priority} that preserves causal structure. Second, the \textbf{Epistemic Reliability Gap} reveals that heuristic proxies (e.g., entropy) often conflate confidence with correctness, demanding \textit{Epistemic Calibration} to ground priority in veridical truth rather than mere likelihood. Finally, the \textbf{Optimization Instability} of static masks against dynamic learning trajectories calls for the development of time-dependent \textit{Priority Schedules}, effectively treating data selection as an optimal control problem rather than a static filtering task.

\section{Future Horizons}
\label{sec:future_work}


\subsection{Redefining SFT: Closing the Generalization Gap}
Current SFT suffers from two major deficits: it is prone to catastrophic forgetting when adapted to new domains, and it generalizes poorly compared to RL because it mimics surface forms rather than reasoning processes. \\
\textbf{The Opportunity:} Token Priority offers a solution to the ``plasticity-stability" dilemma in Continual Learning. Instead of treating all parameters as equally mutable, a priority-weighted objective can act as a dynamic regularization buffer. By identifying and preserving high-value tokens associated with previous skills while selectively updating weights on high-priority tokens for new tasks, we can achieve disjoint capability learning. \\
\textbf{Reframing Generalization:} We posit that the perceived ``generalization gap" between SFT and RL is actually a \textit{selection failure}. RL naturally weights "forking" decisions via reward signals, whereas standard SFT drowns them in trivial tokens. If SFT is re-engineered to explicitly prioritize pivotal reasoning tokens—essentially simulating the credit assignment of RL via static supervision—it may be possible to replicate RL's OOD robustness with SFT's stability \cite{lin2024rho, wang2025beyond, qin2025sstoken}.

\subsection{Importance Sampling for Off-Policy Stability}
The rigid distinction between SFT and Off-Policy RL is largely artificial. We posit that SFT is fundamentally an \textit{Off-Policy} optimization problem: the evolving model $\pi_{\theta}$ must learn from a static ``replay buffer'' (the dataset) generated by a fixed behavior policy $\pi_{\text{data}}$ \citep{ross2010efficient,levine2020offline,wu2025generalization}. Standard SFT ignores the widening \textbf{distribution shift} between the learner $\pi_{\theta}$ and the data $\pi_{\text{data}}$, treating every offline token as an equally valid on-policy signal \citep{yang2024self,kumar2021should}. This leads to the same instability found in Offline RL—overfitting to out-of-distribution trajectories or "exploding" variance when the model drifts \citep{hu2023aligning,liu2025leveraging}. \\
\textbf{The Opportunity:} Token Priority $\Phi(x)$ enables SFT to unlock the stability of rigorous Off-Policy RL. By viewing $\Phi(x)$ as a \textbf{Granular Importance Sampling} estimator (approximating the density ratio $\pi_{\theta}(x)/\pi_{\text{data}}(x)$), we can dynamically modulate SFT gradients. This mechanism allows the model to ``clip'' updates on tokens where the policy divergence is excessive (preventing collapse) while accelerating learning on tokens that lie within the trust region \citep{liutis, ouyang2025token}. This effectively transforms naive SFT into a robust \textbf{Offline RL} process, allowing models to extract safe, high-value signals from massive, noisy offline logs without the complexity of iterative PPO pipelines \citep{Brown2025KLRegularisedQA, tan2025gtpo}.



\section{Alternative Views}
\label{sec:alternative_views}

While the ``Scale is All You Need" hypothesis posits that the gap between the empirical SFT distribution and the ideal alignment manifold is temporary artifacts that vanish with sufficient compute and data \cite{kaplan2020scaling, weiemergent}, We believe that the approach of scaling remains overly simplistic and crude when dealing with inherent structural obstacles. First, uniform optimization induces \textbf{gradient starvation}, where high-frequency surface patterns drown out rare but discriminative reasoning signals, rendering larger models merely more fluent mimics rather than robust reasoners \cite{pezeshki2021gradient, lin2024rho, shen2026training}. Second, the ``LIMA" effect and the risk of ``Model Collapse" contradict the supremacy of raw volume, demonstrating that intelligence relies on a sparse support set of high-density tokens that must be explicitly prioritized to prevent the dilution of signal by noise \cite{zhou2023lima, shumailov2023curse, qin2025sstoken}. Finally, phenomena like \textbf{inverse scaling} reveal that unguided scaling can amplify overfitting to spurious correlations \cite{wei2023inverse, mckenzie2023inverse}, necessitating Token Priority to surgically redirect capacity toward critical decision boundaries rather than minimizing perplexity on the trivial majority.

\section{Conclusion}
We conclude by reiterating that the gap between empirical SFT and ideal alignment cannot be bridged by scale alone. In this paper, we formalized \textbf{Token Priority} as the unifying operator that integrates recent breakthroughs, ranging from constructive noise filtration to corrective unlearning, into a single distribution reshaping process. We challenge the "Scale is All You Need" orthodoxy and advocate for a shift toward dynamic, topology-aware priority schedules to unlock robust general intelligence.

\section*{Impact Statement}

This paper presents a theoretical framework aimed at improving the safety, robustness, and data efficiency of Large Language Model alignment. As a methodological position focused on mitigating hallucinations and optimizing training dynamics, there are no potential negative societal consequences which we feel must be specifically highlighted here.

\nocite{langley00}

\bibliography{example_paper}
\bibliographystyle{icml2026}

\newpage
\appendix
\onecolumn
\section{Detailed Discussion from Limitations to Challenges}
\label{app:challenges}

While the transition from uniform supervision to Token Priority offers a precise mechanism for reshaping the alignment manifold, implementing the priority function $\Phi(x)$ introduces significant structural and optimization bottlenecks. In this section, we analyze three critical limitations in current methodologies—granularity mismatches, epistemic reliability, and dynamic instability—and extrapolate them into the fundamental challenges that future research must address.

\subsection{The Granularity Mismatch: Atomic Optimization vs. Semantic Integrity}
The fundamental premise of Token Priority is that optimization can be performed at the atomic token level. However, language meaning is inherently compositional and non-local.
\begin{itemize}
    \item \textbf{Semantic Fragmentation:} Aggressive \textit{Hard Selection} strategies, such as \textbf{Rho-1} \citep{lin2024rho} or \textbf{ProFit} \citep{liu2026profit}, treat tokens as independent utility units. This risks severing the "connective tissue" of language—prepositions, transition words, or intermediate reasoning steps—that possess low intrinsic information density (low loss/entropy) but are structural prerequisites for coherence. \textbf{T-Shirt} \citep{fu2025t} explicitly identifies that selecting based on individual token scores leads to high-variance neighborhoods, necessitating hierarchical or chunk-wise preservation.
    \item \textbf{The Chain-of-Thought Paradox:} In reasoning tasks, there is a conflict between "answer correctness" and "reasoning fidelity." \textbf{SFTKey} \citep{shi2025rethinking} argues for masking Chain-of-Thought (CoT) tokens to prevent hallucination propagation, effectively setting $\Phi_{CoT} \to 0$. Conversely, \textbf{InfoBottleneck} \citep{lei2025revisiting} and \textbf{SRFT} \citep{fu2025srft} suggest that critical reasoning bottlenecks \textit{must} be emphasized to ensure generalization. This creates a dilemma: purely optimizing for the Answer (via structural priors) may degrade the model's ability to self-correct, while uniformly training on CoT risks memorizing flawed logic.
\end{itemize}

\textbf{The Structural Challenge: From Atomic to Topological Priority.} \\
The limitation of semantic fragmentation implies that $\Phi(x)$ cannot be a scalar function of a single point $x_t$. The future challenge lies in defining \textit{Topological Priority}: constructing $\Phi$ as a function of the semantic dependency graph (e.g., hypergraphs of tokens). How do we mathematically formulate priority such that removing a "low-value" token respects the causal integrity of the high-value tokens that depend on it? Future frameworks may move from independent token scoring to \textit{structure-preserving density estimation}.

\subsection{The Epistemic Reliability Gap: The "Blind Leading the Blind" Problem}
\label{sec:blind}
Priority functions $\Phi(x)$ rely on proxy signals—such as loss, entropy, or reference model likelihood—to estimate data utility. These proxies often fail when the model's epistemic state is misaligned with ground truth.
\begin{itemize}
    \item \textbf{Confidence $\neq$ Correctness:} Methods like \textbf{SelecTKD} \citep{huang2025selectkd} and \textbf{EntroDrop} \citep{wang2025entropy} assume that high confidence (or specific entropy patterns) correlates with data quality. However, \textbf{Faithful FT} \citep{hu2024mitigating} and \textbf{RobustFT} \citep{luo2024robustft} demonstrate that LLMs are prone to "confident hallucinations," where non-factual tokens exhibit low loss. In such cases, standard priority functions inadvertently reinforce errors by assigning high $\Phi$ to hallucinated content.
    \item \textbf{The Reference Bottleneck:} Many Hard Selection methods (\textbf{Rho-1}, \textbf{Anchored SFT}) rely on a reference model $M_{ref}$ to filter noise. This imposes an upper bound: the student model cannot easily surpass the discernment capability of the teacher. If $M_{ref}$ fails to identify a subtle poison token or a complex reasoning step, the priority function effectively blinds the student to these features, validating the need for reference-free methods like \textbf{SelectIT} \citep{liu2024selectit} or \textbf{ssToken} \citep{qin2025sstoken} that rely on intrinsic self-evolution.
\end{itemize}

\textbf{The Verification Challenge: Breaking the Reference Ceiling.} \\
The reliance on imperfect proxies highlights the need for \textit{Epistemic Calibration}. The challenge is to decouple "Statistical Priority" (what the model \textit{thinks} is easy/hard) from "Veridical Priority" (what is actually correct). Future work must develop scalable, reference-free verification mechanisms—potentially leveraging logical consistency checks or multi-view consistency—to ground $\Phi(x)$ in truth rather than likelihood, enabling models to identify priority signals that transcend their current capabilities (Weak-to-Strong Generalization).

\subsection{Optimization Instability: Static Priors in a Dynamic Landscape}
Defining $\Phi(x)$ is not merely a data selection problem but an optimization trajectory problem. Current research highlights the danger of treating priority as a static attribute.
\begin{itemize}
    \item \textbf{Trajectory Mismatch:} \textbf{T3S} \citep{shen2026training} reveals that "token difficulty" is not static; tokens shift from "to-be-learned" to "imitation anchors" (already learned) during training. Static filtration (e.g., pre-computed masks in \textbf{OpenCoder} \citep{huang2025opencoder}) fails to adapt to this drift, potentially wasting compute on learned concepts while under-weighting emerging difficulties. 
    \item \textbf{Unlearning Destabilization:} In Regime II (Correction), methods like \textbf{Forgetting} \citep{ghahrizjani2025forgetting} and \textbf{Ignorance Awareness} \citep{shen2025don} apply negative gradients ($\Phi < 0$) to erase harmful behaviors. However, \textbf{Anchored SFT} \citep{zhu2025anchored} warns that unconstrained reweighting (or negative updates) causes distributional drift, degrading general capabilities. The challenge lies in applying surgical penalties to specific "hallucination modes" without collapsing the model's broader knowledge representation.
\end{itemize}

\textbf{The Control Challenge: Optimal Scheduling of Priority.} \\
The static nature of current priority masks fails to capture the learning dynamics. The grand challenge is to formulate $\Phi(x, t)$ as a time-dependent control policy. Similar to learning rate schedules, we need \textit{Priority Schedules}: when should a token transition from being "Constructive" ($\Phi > 0$, to be learned) to "Neutral" ($\Phi \approx 0$, learned/ignored) or "Corrective" ($\Phi < 0$, unlearned)? Solving this requires a theoretical framework for the "Optimal Transport" of probability mass over training time, moving beyond static data selection to dynamic curriculum control.

\end{document}